\documentclass{article}

\PassOptionsToPackage{round, sort&compress}{natbib}

\usepackage[preprint]{nips_2018}

\usepackage[utf8]{inputenc} 
\usepackage[T1]{fontenc}    
\usepackage{hyperref}       
\hypersetup{colorlinks=true,
    linkcolor=[rgb]{0,0,1},
    citecolor=[rgb]{0,0,1},
    urlcolor=[rgb]{0,0,1}}
\usepackage{url}            
\usepackage{booktabs}       
\usepackage{amsfonts}       
\usepackage{nicefrac}       
\usepackage{microtype}      
\usepackage{graphicx}

\usepackage[draft=true]{minted}
\usepackage{authblk}
\usepackage{caption}

\usepackage{amsmath}
\usepackage{amssymb}
\usepackage{subfigure}
\usepackage{enumitem}
\usepackage{pdfpages}
\usepackage{tabularray}

\title{Orion-14B: Open-source Multilingual \\Large Language Models}
\author{OrionStar Inc.\thanks{Authors are listed in Appendix A.}}
\date{Jan 2024}

\begin{document}

\maketitle

\begin{abstract}
In this study, we introduce Orion-14B, a collection of multilingual large language models with 14 billion parameters. We utilize a data scheduling approach to train a foundational model on a diverse corpus of 2.5 trillion tokens, sourced from texts in English, Chinese, Japanese, Korean, and other languages. Additionally, we fine-tuned a series of models tailored for conversational applications and other specific use cases. Our evaluation results demonstrate that Orion-14B achieves state-of-the-art performance across a broad spectrum of tasks. We make the Orion-14B model family and its associated code publicly accessible\footnote[1]{\url{https://github.com/OrionStarAI/Orion}}, aiming to inspire future research and practical applications in the field.
\end{abstract}

\section{Introduction}
Three hundreds years ago, Gottfried Wilhelm Leibniz's insightful declaration that "Language is the mirror of the mind" profoundly resonates in the contemporary exploration of language. This thought provides a philosophical foundation for understanding the intricate relationship between language and intelligence. In the 20th century, language modeling (LM) became a fundamental approach in artificial intelligence, forming the cornerstone of natural language processing (NLP). The goal of language modeling is to learn the probability distribution of word sequences. Desipite its simple modeling procedure, it encapsulates substantial information about languages. Given that a language contains $N$ words, the potential number of word sequences of the length of $L$ is $N^L$. However, the actual number of sentences commonly used in the language is far fewer than $N^L$. This discrepancy highlights how language models effectively encode linguistic information.

Traditionally, statistical methods were employed to model word frequency in a language. Among these, the $N$-gram model has been widely adopted, determining the probability of a word based on the previous $N-1$ words. Though straightforward and efficient, the method suffers from the data sparsity problem. With the advancement of neural networks, a paradigm shift occurred towards employing neural networks for language modeling. There are many variations of neural language models, such as multi-layer perceptron (MLP) \citep{bengio2000neural}, recurrent neural networks (RNN) \citep{mikolov2010recurrent,yao2013recurrent}, and transformer \citep{vaswani2017attention,devlin2019bert}.

In recent years, the increase of model sizes and the scale of training data have significantly boosted the capability of language models \citep{Peters:2018,radford2018improving,devlin2019bert}. Large language models (LLMs) have exhibited remarkable promise in many traditional NLP tasks, such as dialogue system, machine translation, information retrieval. Moreover, LLMs have proven adept at complex tasks such as reasoning, code generation, creative writing. These advancements have inspired both the academic and industrial sectors to further investigate the underlying principles and potential applications of LLMs.

The launch of ChatGPT/GPT-3.5 \citep{chatgpt} in 2022 captured tremendous attention from the public, pushing the boundaries of what AI can achieve and motivating researchers and engineers to delve deeper into their potential. While GPT-3.5 and its successor, GPT-4 \citep{gpt4}, are prime examples of LLMs, their specific model architectures and training methodologies remain undisclosed. In contrast, Meta's release of LLaMA \citep{LLaMA} and LLaMA 2 \citep{LLaMA2} have established a widely-recognized LLM architecture within the open-source community, with numerous libraries adopting these models. Despite LLaMA's impressive performance, its primary focus on English limits its applicability to other languages. Recently, there has been a surge in the release of multilingual LLMs such as ChatGLM \citep{Chatglm3}, Baichuan \citep{baichuan1,baichuan2}, Qwen \citep{Qwen}, InternLM \citep{2023internlm}, XVERSE \citep{XVERSE-13B}, Skywork \citep{skywork} and Yi \citep{Yi}. These models, trained on mainly English and Chinese datasets, have shown promising results in tasks involving both English and Chinese. Additionally, there has been a growing trend of LLMs specifically designed to enhance performance in other languages, such as Japanese \citep{PLaMo2023Introducing,elyzaLLaMA2023,weblab-10b,JapaneseStableLMBaseAlpha7B}and Korean \citep{KoGPT,Polyglot-Ko}.

In this technical report, we present Orion-14B, a family of multilingual language models with 14 billion parameters. The foundation model is trained on a diverse dataset of 2.5 trillion (2.5T) tokens, containing languages such as English, Chinese, Japanese, Korean, and others. It has demonstrated superior performance across a broad spectrum of tasks in multilingual settings. 

We also provides a series of fine-tuned models built upon the foundation model, each trained to different focuses such as conversation, long-context text handling, quantization, and specific application requirements.

The remainder of this report describes our data preparation (Section \ref{sec:data}), pretraining methodology (Section \ref{sec:pretrain}), fine-tuning methodology (Section \ref{sec:finetune}), evaluation results (Section \ref{sec:eval}), extension works (Section \ref{sec:ext}), and conclusions (Section \ref{sec:conclusion}).

\section{Data}
\label{sec:data}

In the training framework of LLMs, the role of data is crucial in determining the efficacy and performance of these models. Effective data processing for pretraining is essential for achieving the desired outcomes. This involves acquiring data from diverse sources, ensuring the high quality of the data through thorough filtering, and removing any redundant information. This section will discuss these processes in detail, outlining the necessary steps to prepare and refine data to suit the stringent requirements of LLM training.

\subsection{Data Source}
\label{subsec:subection2.1}
Pretraining of LLM needs tremendous amounts of data. \cite{hoffmann2022training} offered guildlines regarding the optimal quantity of training data for models of varying sizes. For example, an LLM with 10 billion parameters requires 205 billion tokens for pretraining. However, recent work \citep{LLaMA2,baichuan2,skywork} in training 10 billion parameter models have utilized 2.5 to 3 trillion tokens, substantially exceeding the recommended data volume. These efforts have yielded notably impressive results, demonstrating the efficacy of training with significantly larger datasets than those proposed in the aforementioned study.

In order to obtain such a large amount of data, it is imperative to collect data from multitude of sources with diversity and high quality. Our dataset incorporates texts in multiple languages, with English and Chinese being predominant, constituting over 90\% of the entire dataset. Particular efforts are also made to include Japanese and Korean texts, which account for more than 5\% of the dataset. The remaining portion comprises texts in various other languages, such as Spanish, French, German, Arabic, and more.

\begin{figure}[ht]
\centering
\includegraphics[width=0.8\textwidth]{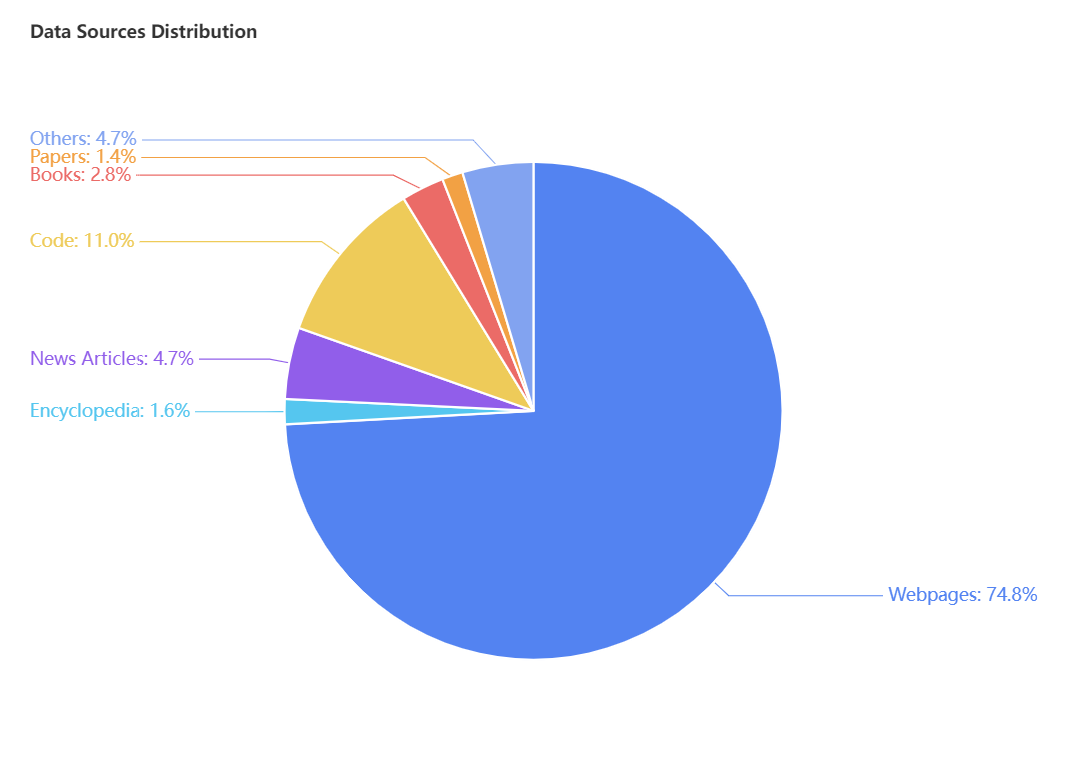}
\caption{Data sources distribution.}
\label{fig:data_source}
\end{figure}

In terms of content and style, the dataset primarily composes of written language, with spoken language constituting only a minor portion. The dataset spans a broad spectrum of topics, including web pages, news articles, encyclopedic entries, books, source code, and academic publications, among others. The diversity within the dataset is a crucial factor in achieving superior performance across a range of tasks. The detailed distribution of the data sources is shown in Fig. \ref{fig:data_source}. We believe that different types of corpora exert varying influences on the model training process; for instance, some may be more effective to language understanding, while others better facilitate knowledge reasoning. Unlike many existing studies that typically employ random shuffling of training examples, we strategically feeds the model with varied data sources across different training stages. We believe this method leads to more efficient data usage. The details of this approach will be elaborated in Section \ref{sec:pretrain}.

\subsection{Data Quality}
Data quality is essential in the training of LLMs.  To assure high-quality data, we have implemented a series of measures for data filtering, detailed as follows:

\begin{itemize}
    \item \textbf{Text normalization}: The datasets contain a large number of texts from various sources, such as web pages and ebooks. These texts are often accompanied by HTML, special characters, or other format tags, which are not useful for LLM training. We employ a series of tools, such as regular expressions and format parsers, to effectively eliminate them.
    
    \item \textbf{Harmful content removal}: The Internet contains harmful and spam content. Our approach to mitigate this involves a two-stage process: the initial stage utilizes keywords and regular expressions matching, followed by a deep learning-based model designed to identify and remove such content. It is important to note that entirely eliminating harmful text from the training dataset could lead to a scenario where the trained model lacks the ability to identify and appropriately response to harmful information \citep{LLaMA2}. Therefore, we intentionally retain a minimal amount of harmful text in the dataset. This approach ensures that the model remains capable of recognizing and effectively addressing such content.

    \item \textbf{Personal information removal}: Some of the text data includes personal details like names, phone numbers, and addresses. We utilize rule-based methods for detection and either substitute these with placeholders or remove them entirely.

    \item \textbf{Quality filtering}: This part is crucial in data processing. We first apply a set of rules to filter the data, such as removing texts with excessive repetition. Additionally, we use $N$-gram perplexity models to exclude texts with anomalously high perplexity. Lastly, our proprietary data quality models are employed to select high-quality data. We emphasize that while high quality is essential for LLM training, achieving a balance between quality and quantity of training data is a non-trivial task. Our models are optimized to retain as much data as possible while maintaining high data quality, which is one of the key factors in the successful training of LLMs.
\end{itemize}

\subsection{Deduplication}
Given that the training data for LLMs is sourced from a variety of origins, there is a significant likelihood of encountering duplicate data. Duplicate data can detrimentally affect the training process, potentially leading to a model biased towards certain data sources and a decline in performance \citep{lee2021deduplicating,nunes2023dothash,penedo2023refinedweb}. To address this, we develop a deduplication procedure to eliminate redundant data.

In this process, we extract key words and phrases from each document and compute their corresponding embedding vectors and SimHash vectors \citep{indyk1998approximate,charikar2002similarity}. These vectors are then compared to those in our database. If a vector in the database shows similarity within a certain threshold, the document is considered a duplicate and is subsequently discarded. 

Importantly, we note that while LLMs have shown significant advancements in numerous NLP tasks, some studies \citep{yang2023rethinking,golchin2023time,skywork} indicate that part of this improvement might be attributed to unintentional inclusion of evaluation data in the training datasets, potentially leading to overestimated results. To address this, we adopt our deduplication approach for all evaluation datasets to prevent the pretraining dataset from containing texts in the evaluation sets, thereby enhancing the integrity and reliability of our model's evaluation results. We will further discuss the data contamination in detail in Section \ref{subsec:data_contamination}.

\section{Pretraining}
\label{sec:pretrain}


\subsection{Tokenizer}
\label{subsec:subection3.1}
A tokenizer is a basic component of an LLM, which need to represent the text distribution in the language while maintaining an favorable vocabulary size for training. For a multilingual tokenizer, statistical methods are typically employed to generate word-level or subword-level tokens from texts in multiple languages. We utilize the byte-pair encoding (BPE) algorithm \citep{shibata1999byte}, implemented via SentencePiece \citep{kudo2018sentencepiece}. Our configuration ensures a character coverage of 99.99\%, with rare characters defaulting to UTF-8 bytes. To build a diverse corpus and align with our training data distribution, we curate a broad spectrum of text types from our training corpus. This includes English, Simplified Chinese, Traditional Chinese, Japanese, Korean, a few other languages, as well as rare characters. In Table \ref{tab:tokenizer}, we provide a detailed comparison of our tokenizer with other open-source tokenizers. This comparison includes vocabulary size and compression ratio (CR), the latter calculated by the ratio of the size of the original data to the size of the tokenized data.

\begin{table}[h]
\centering
\caption{Tokenizer comparison with other open-source LLMs. We compare vocabulary sizes and compression ratios for simpifiled Chinese (zh\_cn), tranditional Chinese (zh\_cn), and English, respectively.}
\label{tab:tokenizer}
\begin{tabular}{lcccc}
\hline
Model & Vocab Size & CR (zh\_cn) & CR (zh\_tw) & CR (en)  \\
\hline
LLaMA2 & 32,000 & 1.377  & 1.589 & 1.153 \\
Yi & 64,000 & 0.606 & 0.785 & 1.084 \\
Baichuan2 & 125,696 & 0.554  & 0.783 & 1.077 \\
ChatGLM3 & 65,024 & 0.582  & 0.703 & 1.081 \\
Skywork  & 65,519 & 0.672 & 0.846 & 1.153 \\
\hline
Orion-14B & 84,608 & 0.549 & 0.656 & 1.067 \\
\hline
\end{tabular}
\end{table}

\subsection{Architecture}
\label{subsec:subection3.2}
Given that LLaMA2 has achieved superior performance, its architecture has been widely adopted by many open-source LLM. In our approach, we adhere to the LLaMA2 architecture while implementing several modifications. These include extending the number of tokens to 84,608 and enlarging the dimensions of the feed-forward network (FFN) to 15,360. We employ rotary positional embeddings (RoPE) \citep{su2021roformer} for positional encoding to accommodate context lengths of up to 4096 tokens. The model uses 40 transformer layers with 40 attention heads each. The total parameter of the model is 14.4 billion, slightly exceeding that of LLaMA2-13B. Detailed model parameters is provided in Table \ref{tab:arch}.

\begin{table}[h]
\label{token}
\centering
\caption{A comparison of model architecture. The table shows comparison of our model and several open-source model with similar model size.}
\label{tab:arch}
\begin{tabular}{lcccccc}
\hline
Model & seq\_len & position embedding & hidden size & FFN size & \# layers & \# heads \\
\hline
LLaMA2-13B & 4096 & RoPE & 5,120 & 13,824 & 40 & 40 \\
Skywork-13B & 4096 & RoPE & 5,120 & 12,288 & 52 & 36 \\
Baichuan2-13B & 4096 & AliBi & 5,120 & 13,696 & 40 & 40 \\
Qwen-14B & 2048 & RoPE & 5,120 & 13,696 & 40 & 40 \\
InternLM-20B & 4096 & RoPE & 5,120 & 13,824 & 60 & 40 \\
\hline
Orion-14B & 4096 & RoPE & 5,120 & 15,360 & 40 & 40 \\
\hline
\end{tabular}
\end{table}

\subsection{Infrastructure}
\label{subsec:subection3.3}

For the training of Orion-14B, we employed Megatron-LM \citep{shoeybi2020megatronlm} on a cluster comprising 11 servers, each equipped with 8 NVIDIA H800 GPUs. To optimize training efficiency, we integrated FlashAttention2 \citep{dao2023flashattention2} and APEX \citep{apex} into Megatron-LM framework, achieving a training speed of 4,000-5,000 tokens/GPU/second.

\subsection{Training Process}
\label{subsec:subection3.4}
To train Orion-14B, we initiate the model training with a learning rate warm-up stage spanning 2000 iterations, during which we linearly increase the learning rate to the maximal learning rate of 3e-4. We then apply a cosine schedule to gradually decrease the learning rate to 3e-5 throughout the training processing. We employ the AdamW \citep{loshchilov2018fixing} optimizer, setting $\beta_1$ to 0.9 and $\beta_2$ to 0.95, respectively. In addition, we apply a weigh decay factor of 0.1 and enforce a gradient clipping threshold of 1.0 to ensure the stability of the training process. The model is trained using BF16/FP32 mixed precision, with a batch size of 1408, corresponding to approximately 5.7 million tokens per step.

\subsection{Data Scheduling}
Training large language models requires hundreds of billions to trillions of tokens. It is an interesting area to explore scaling laws in LLM training and literature from \cite{kaplan2020scaling} through \cite{hoffmann2022training} to \cite{LLaMA2} suggests that model training tends to favor an increase in the number of tokens over model sizes. We use a 2.5T token training dataset for our 14B parameter model, aiming a balance between computational efficiency and cost.

On the other side, while numerous theoretical and empirical studies have examined the interplay between model size and training data volume, there is no universally accepted methodology for scheduling training data. Considering that humans acquire knowledge in a deliberate order \citep{evanson2023language}, it is plausible that language models might also benefit from a structured learning order when processing training data. Curriculum learning \citep{bengio2009curriculum} has been suggested as a method to organize the learning process by progressively increasing the complexity of the training data. However, most prior studies have concentrated on sample-level data and smaller datasets. \cite{chen2023skill} employed a skills-based framework for training data selection and continuous pretraining with a 3B-parameter language model. This approach achieved greater accuracy compared to the baseline method of uniform data source sampling, suggesting the potential efficacy of strategic data scheduling.

In training the Orion-14B model, we intentionally develop a data scheduling strategy that organizes training data to incrementally increase its complexity. We divide the training data into several stages based on the data sources and their complexity. These stages are differentiated by the mix ratios of data sources. Initial stages primarily include data with common knowledge, such as web pages and news articles. In the subsequent stages, we gradually augment the proportion of data containing more complex knowledge, including textbooks, academic papers, and source code. Additionally, the linguistic diversity of the training data is expanded progressively from English and Chinese to Japanese and Korean. The brief structure of our training data schedule is depicted in Table \ref{tab:schedule}.

\begin{table}[h]
\label{token}
\centering
\caption{Training data schedule for Orion-14B. Primary data sources and languages refer to data that totally account for more than 90\% of the whole training data in each stage.}
\label{tab:schedule}
\begin{tabular}{llp{5cm}l}
\hline
Stages & Tokens (B) & Primary data sources & Primary languages \\
\hline
Early stages & 0 \textasciitilde 600 & web pages, news articles & English, Chinese  \\
Middle stages & 600 \textasciitilde 1100 & web pages, news articles, textbooks, academic papers & English, Chinese, Others \\
Final stages & 1100 \textasciitilde 2000 &  web pages, news articles, textbooks, academic papers, source code & English, Chinese, Others\\
\hline
\end{tabular}
\end{table}

\begin{figure}[htbp]
\centering
\includegraphics[width=0.8\textwidth]{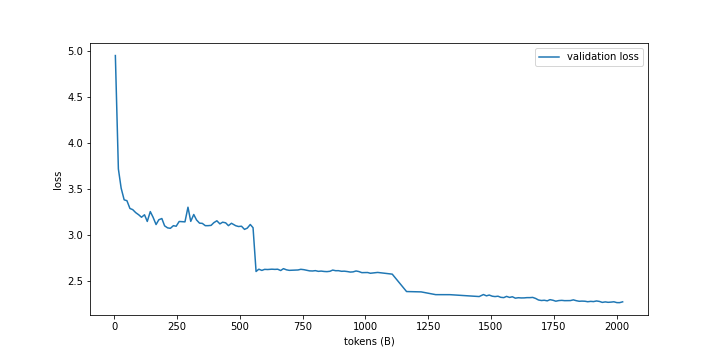}
\caption{Validation loss during the training process. The validation set consists of 5,000 documents including a diverse collection of English and Chinese texts sourced from a variety of data sources.}
\label{fig:valid_loss}
\end{figure}

To assess the effectiveness of the data scheduling approach, we monitor the loss on a validation set throughout the training process. This validation set consists of 5,000 documents, each unseen in the training set. It includes a diverse collection of English and Chinese texts sourced from a variety of data sources. As shown in Fig. \ref{fig:valid_loss}, there are significant reduction in validation loss aligning with shifts in the training data distribution at 600B and 1,100B tokens. Additionally, the validation loss exhibits initial fluctuations, stabilizing progressively with continued training. This trend indicates that the model increasingly adapts to the diversity of data types as training progresses.

To our knowledge, the training of most prior LLMs utilized fully shuffling the training data, which was then fed into the model in a random sequence. Orion-14B is the first LLM trained with a specific data scheduling strategy. The evaluation results indicate that this model demonstrates impressive performance in language understanding tasks at its early stages and rapidly enhances its capabilities in reasoning and academic tasks in later stages, aligning with our data scheduling policy. Notably, Orion-14B, trained on 2.5T tokens, achieves comparable performance to other open-source models trained on 2.6T to 3T tokens, thereby illustrating the efficiency of our data utilization approach. 

\section{Fine-tuning}
\label{sec:finetune}
During the pretraining stage, an LLM is trained to predict the next token at each step. However, in many applications, the model needs to generate a desired response to a given prompt. Thus, in the next stage, LLMs typically undergo further fine-tuning using supervised learning, where the training data consists of paired input and output text sequences. Further, to enhance the quality and safety, approaches like Reinforcement Learning from Human Feedback (RLHF) \citep{christiano2017deep,ouyang2022training} or Direct Preference Optimization (DPO) \citep{rafailov2023direct} can be employed. In this work, our primary focus is on the supervised fine-tuning (SFT) stage, leaving RLHF and DPO for future exploration.

\subsection{SFT Data}
High-quality, diverse data has been proven to be crucial to supervised fine-tuning in previous studies \citep{LLaMA2,zhou2023lima}. To construct our SFT training data, we draw from two primary sources: a human-labeled dataset and an open-source filtered dataset.

For a high-quality human-labeled dataset, we assemble a team of expert annotators who spend weeks creating precisely annotated data. To ensure data quality, all annotators adhere to three key principles—helpfulness, truthfulness, and harmlessness—as outlined in InstructGPT \citep{ouyang2022training} and LLaMA2 \citep{LLaMA2}. In total, we produce approximately 220,000 human-labeled SFT data entries.

While the human-labeled dataset is of high quality, its volume is insufficient for training a high-performance LLM. Therefore, we also construct a large-scale, open-source filtered dataset. The original SFT data includes collections from various open-source datasets, such as COIG \citep{zhang2023chinese}, WildChat \citep{zhao2023wildchat}, OpenOrca \citep{OpenOrca}, and UltraChat \citep{ding2023enhancing}. Given the variable quality and occasional presence of inappropriate content in these open-source datasets, we implement a cleaning process inspired by methods from Platypus \citep{lee2023platypus} and MoDS \citep{du2023mods}, comprising the following steps:

\begin{itemize}
\item Rule-based filtering: We use regular expressions and keywords for simple filtering to remove personal information, temporal-sensitive data, etc.
\item Quality filtering: A large NLP model scores the data quality on a scale from 1 to 10, retaining only data with a score of 7 or higher.
\item Semantic deduplication: Text embeddings are used for semantic deduplication, considering texts with a similarity greater than 0.98 as duplicates.
\end{itemize}

Using this approach, we construct an open-source filtered dataset of 630,000 samples. Combined with the human-labeled data, we assemble an SFT dataset of 850,000 training pairs for supervised fine-tuning.

\subsection{Training details}
To fine-tune a pretrained LLM, we prepend \texttt{<human>} and \texttt{<assistant>} as headers to the prompt text and the response text, respectively. The training process employs the AdamW optimizer, with hyperparameters configured as follows: $\beta_1$ is set to 0.9, $\beta_2$ to 0.95, and $\epsilon$ to $1e-8$. We limit the sequence length to 4096 and use a batch size of 128. Our training regimen spanned three epochs, involving over 500k samples; the learning rate was incrementally increased over the first 1,500 steps to a maximum of $1e-5$. To prevent overfitting, we apply a weight decay of 0.1, a dropout rate of 0.1, and a gradient clipping threshold of 1.0.


\section{Evaluation}
\label{sec:eval}
\subsection{Standard Evaluation}
To effectively evaluate the LLM, we categorize the standard evaluation sets into the examinations and professional knowledge, and language understanding and common knowledge. We select the most common evaluation sets in each category as follows:

\textbf{Professional Knowledge and Reasoning}
\begin{itemize}
    \item \textbf{C-Eval} \citep{Ceval}: A comprehensive Chinese evaluation benchmark consisting of more than 10,000 multi-choice questions.

    \item \textbf{CMMLU} \citep{cmmlu}: A general evaluation benchmark specifically designed to evaluate the knowledge and reasoning abilities of LLMs within the context of Chinese language and culture.

    \item \textbf{MMLU} \citep{mmlu}: A widely used benchmark designed to measure knowledge acquired during pretraining by evaluating models.
    

    \item \textbf{AGIEval} \citep{zhong2023agieval}: A human-centric benchmark crafted to assess the general capabilities of foundation models in tasks aligned with human cognition and problem-solving.

    \item \textbf{Gaokao} \citep{Zhang2023EvaluatingTP}: A dataset leverages questions from China's national college entrance examination to test LLMs. 

    \item \textbf{BBH} \citep{suzgun2022challenging}: A challenging subset of the Big-Bench suite, covering a wide array of themes, such as linguistics, mathematics, common sense reasoning, biology, physics, software development, and more.
    
\end{itemize}

\textbf{Language Understanding and Common Knowledge}
\begin{itemize}
    \item \textbf{RACE} \citep{lai2017large}: A comprehensive reading comprehension dataset comprising over 28,000 passages and nearly 100,000 questions. It contains reading and comprehension materials for both middle school (middle) and high school (high) academic levels.
    
    \item \textbf{HellaSwag} \citep{hellaswag}: A challenge dataset for evaluating commonsense language inference that is particularly difficult for state-of-the-art models.
    
    \item \textbf{PIQA} \citep{piqa}: A dataset introducing the task of physical commonsense reasoning and a corresponding benchmark dataset.
    \item \textbf{Lambada} \citep{lambada}: A collection of narrative passages where human subjects can guess the last word if exposed to the whole passage, but not if they only see the last sentence preceding the target word.
    
    \item \textbf{WSC} \citep{WSC}: A pronoun disambiguation task, which requires determining the noun that the pronoun refers to according to the context.
\end{itemize}


For comparison, we select the most popular LLMs with a parameter range of 10-20 billion: LLaMA 2-13B \citep{LLaMA2}, Skywork-13B \citep{skywork}, Baichuan 2-13B \citep{baichuan2}, Qwen-14B \citep{Qwen}, InternLM \citep{2023internlm}.

To ensure consistent comparisons, we employ open-source LLM evaluation frameworks such as OpenCompass \citep{opencompass} and LM-Eval-Harness \citep{eval-harness} for a unified performance evaluation of LLMs. For the models we compared, we refer to the published scores from OpenCompass or their official reports.

\begin{table}[h]
\centering
\caption{LLM evaluation results on examination and professional knowledge. Bold font denotes the best score in each category, a convention followed in all subsequent tables throughout this paper.}
\label{exp-exam}
\begin{tabular}{l|cccccc}
\hline
Model & C-Eval & CMMLU & MMLU  & AGIEval& Gaokao & BBH\\ 
\hline
LLaMA 2-13B &  41.4&  38.4&   55.0&  30.9& 18.2 & 45.6\\
Skywork-13B &  59.1&  61.4&   62.7&  43.6&  56.1& 48.3\\
Baichuan 2-13B &  59.0&  61.3&   59.5&  37.4& 45.6 & 49.0\\
Qwen-14B &  71.7&  70.2&   67.9&  51.9& \textbf{62.5} & 53.7\\
InternLM-20B &  58.8&  59.0&   62.1&  44.6& 45.5 & 52.5\\
\hline
\textbf{Orion-14B}  &  \textbf{72.9}&  \textbf{70.6}&   \textbf{69.9}& \textbf{54.7}& 62.1& \textbf{56.5}\\
\hline
\end{tabular}
\end{table}

The evaluation results in Table \ref{exp-exam} highlight Orion-14B's superior performance across various examinations and professional knowledge evaluation sets, compared to other LLMs. Orion-14B achieves the highest scores in C-Eval, CMMLU, MMLU, AGIEval, and BBH, indicating its strong capabilities in understanding and reasoning within professional contexts. While it excels in most benchmarks, it is slightly outperformed by Qwen-14B in the Gaokao evaluation. These results position Orion-14B as a highly competitive and robust model for complex and professional tasks.

\begin{table}[h]
\centering
\caption{LLM evaluation results on language understanding and common knowledge.}
\label{exp-nlu}
\begin{tabular}{l|ccccccc}
\hline
Model &RACE-middle & RACE-high & HellaSwag & PIQA  &  Lambada& WSC \\
\hline
LLaMA 2-13B &  63.0& 58.9 & 77.5  & 79.8 &  76.5& 66.3\\
Skywork-13B & 87.6& 84.1& 73.7 & 78.3 &  71.8& 66.3\\
Baichuan 2-13B & 68.9& 67.2 &70.8  &78.1  &  74.1& 65.4\\
Qwen-14B & 93.0& 90.3& \textbf{80.2}  & 79.8 & 71.4& 66.3\\
InternLM-20B & 86.4& 83.3& 78.1& \textbf{80.3} & 71.8& 68.3\\
\hline
\textbf{Orion-14B} &  \textbf{93.2}& \textbf{91.3}& 78.5& 79.5&  \textbf{78.8}& \textbf{70.2}\\
\hline
\end{tabular}
\end{table}

As shown in Table \ref{exp-nlu}, Orion-14B showcases robust performance in language understanding and common knowledge tasks, outperforming other models in RACE (mid and high), Lambada, and WSC benchmarks, highlighting its exceptional comprehension and reasoning abilities. However, for HellaSwag, PIQA, and WSC tasks, it is slightly outperformed by Qwen-14B and InternLM-20B. Overall, the results indicate Orion-14B's strong capabilities across a spectrum of natural language understanding benchmarks.

For a comprehensive evaluation, we also utilize all test sets used in OpenCompass leaderboard \citep{opencompass} to assess performance. In OpenCompass leaderboard, the evaluation sets are organized into five categories. The summarized results for each category are shown in Table \ref{tab:oc1}, where Orion-14B leads with an average score of 64.4\%. Notably, it outperforms other models across four categories, including Examination, Language, Understanding, and Reasoning, indicating its excellent analytical and problem-solving abilities. These results demonstrate Orion-14B's robust capabilities in a wide range of cognitive and language tasks. Detailed results for each testset are included in the Appendix \ref{sec:opencompass}.

\begin{table}[h]
\centering
\caption{LLM evaluation results of OpenCompass testsets}
\label{tab:oc1}
\begin{tabular}{l|c|ccccc}
\hline
Model & Average & Examination & Language & Knowledge & Understanding  &  Reasoning \\
\hline
LLaMA 2-13B & 47.3&  45.2& 47.0& 58.3  & 50.9 & 43.6 \\
Skywork-13B & 53.6& 61.1& 51.3 & 52.7 & 64.5 & 45.2\\
Baichuan 2-13B &49.4& 51.8& 47.5 & 48.9 & 58.1 & 44.2\\
Qwen-14B & 62.4& 71.3& 52.7 & 56.1  & 68.8 & 60.1\\
InternLM-20B &59.4 & 62.5& 55.0& \textbf{60.1}& 67.3& 54.9 \\
\hline
\textbf{Orion-14B} &\textbf{64.3} &  \textbf{71.4}& \textbf{55.0}& 60.0&  \textbf{71.9}& \textbf{61.6}\\
\hline
\end{tabular}
\end{table}

Note that, evaluation scores are not the definitive standard for assessing an LLM. Given the vast amount of training data, there is a high likelihood that the dataset includes elements of the evaluation set. To avoid this, we purposely deduplicate the evaluation datasets from our pretraining corpus, thereby ensuring that our model's performance genuinely reflects its capabilities. Overlooking this critical step could lead to training a model that is overfitted to the evaluation set, resulting in artificially high scores. We will delve into this topic more deeply in Section \ref{subsec:data_contamination}.


\subsection{Multilingual}
In our training approach, while the majority of the data is in English and Chinese, we also incorporate additional languages to enhance multilingual performance. Notably, Japanese and Korean texts are specifically added after surpassing 600B tokens in the training process. The total amounts of Japanese and Korean texts are approximately 100B and 50B tokens, respectively. Despite the lower quantity of Japanese and Korean tokens compared to English and Chinese, the model exhibits superior performance in these languages. This indicates a significant transfer of knowledge from the more dominant languages during the training of the LLM.

To assess the model's multilingual capabilities, we benchmark it against other models trained on English+Japanese \citep{PLaMo2023Introducing,weblab-10b,elyzaLLaMA2023,JapaneseStableLMBaseAlpha7B}, English+Korean \citep{KoGPT,ko2023technical}, or multilingual datasets \citep{LLaMA2,baichuan2,Qwen,Yi}. We employ the datasets from \cite{eval-harness} and \cite{kobest} for evaluation of Japanese and Korean, respectively.

\begin{table}[h]
\centering
\caption{Comparison of LLM performances on Japanese testsets. The header of each column stands for Japanese CommonsenseQA, Japanese NLI, MARC in Japanese, Japanese SQUAD, Japanese QKET\_v2, XLSUM in Japanese, XWinograd in Japanese, MGSM, respectively.}
\label{jaeval}
\begin{tabular}{l|c|ccccccccccc}
\hline
Model & \textbf{Average} & JCQA& JNLI& MARC& JSQD& JQK& XLS& XWN& MGSM\\
\hline
PLaMo-13B & 52.3& 56.7& 42.8& 95.8&  70.6& 71.0&  8.70&  70.5&  2.40&\\
WebLab-10B &50.7& 66.6& 53.7& 82.1&  62.9& 56.2& 10.0&  72.0&  2.40&\\
ELYZA-jp-7B &48.8&  71.7& 25.3& 86.6& 70.8& 64.1&  2.50& 62.1 & 7.20& \\
StableLM-jp-7B &51.1& 33.4& 43.3& \textbf{96.7}& 70.6& 78.1&  10.7&  72.8& 2.80&\\
\hline
LLaMA 2-13B &46.3& 75.0& 47.6& 38.8 & 76.1& 67.7& 18.1 &  63.2 & 10.4&\\
Baichuan 2-13B&57.1& 73.7& 31.3& 91.6& 80.5& 63.3& 18.6& 72.2& 25.2 & \\
Qwen-14B& 65.8& 85.9& 60.7& 97.0& 83.3& 71.8& 18.8& 70.6& 38.0& \\
Yi-34B& 67.1& 83.8& 61.2& 95.2& \textbf{86.1}& 78.5& \textbf{27.2}& 69.2& 35.2& \\
\hline
\textbf{Orion-14B} & \textbf{69.1}& \textbf{88.2}& \textbf{75.8}& 94.1& 75.7&  \textbf{85.1}& 17.3&  \textbf{78.8}& \textbf{38.0}& \\
\hline
\end{tabular}
\end{table}

\begin{table}[h]
\centering
\caption{Comparison of LLM performances on Korean testsets. $n=0$ and $n=5$ stand for $n$-shot prompts used in the evaluation. The testsets are originally in English and have been translated to Korean by \cite{kobest}.}
\label{koeval}
\begin{tabular}{l|cc|cccccccc}
\hline
& \multicolumn{2}{c}{\textbf{Average}} & \multicolumn{2}{c}{HellaSwag}& \multicolumn{2}{c}{COPA}& \multicolumn{2}{c}{BooIQ} & \multicolumn{2}{c}{SentiNeg}\\
Model & n=0 & n=5& n=0 & n=5& n=0 &  n=5 & n=0 & n=5 &n=0 & n=5\\
\hline
KoGPT& 53.0& 70.1& 55.9& 58.3& 73.5& 72.9& 45.1& 59.8& 37.5& 89.4 \\
Polyglot-ko-13B& 69.6& 73.7& \textbf{59.5}& \textbf{63.1}& \textbf{79.4}& \textbf{81.1}& 48.2& 60.4& 91.2& 90.2 \\
\hline
LLaMA 2-13B& 46.7& 63.7& 41.3& 44.0& 59.3& 63.8& 34.9& 73.8& 51.5& 73.4 \\
Baichuan 2-13B& 52.1& 58.7& 39.2& 39.6& 60.6& 60.6& 58.4& 61.5& 50.3& 72.9 \\
Qwen-14B & 53.8& 73.7& 45.3& 46.8& 64.9& 68.9& 33.4& 83.5& 71.5& 95.7\\
Yi-34B& 54.2& 72.1& 44.6& 44.7& 58.0& 60.6& 65.9& 90.2& 48.3& 92.9 \\
\hline
\textbf{Orion-14B} & \textbf{74.5}& \textbf{79.6} & 47.0& 49.6& 77.7& 79.4& \textbf{81.6}& \textbf{90.7}& \textbf{92.4}&\textbf{98.7} \\
\hline
\end{tabular}
\end{table}

\begin{table}[h]
\centering
\caption{Multilingual evaluation.}
\label{tab:multilingual}
\begin{tabular}{l|c|cccccc}
\hline
Model &Train Lang  &Japanese &Korean  &Chinese & English \\
\hline
PLaMo-13B& En,Jp  & 52.3  & * & * & * \\
Weblab-10B& En,Jp  & 50.7  & * & * & * \\
ELYZA-jp-7B & En,Jp  & 48.8 & * & * & *  \\
StableLM-jp-7B  &En,Jp  & 51.1  & * &*  &*  \\
\hline
KoGPT-6B & En,Ko& * & 70.1 & * & * \\
Polyglot-ko-13B & En,Ko & * & 70.7 & * & * \\
\hline
Baichuan2-13B & Multi& 57.1 & 58.7 & 50.8 & 57.1  \\
Qwen-14B & Multi & 65.8 & 73.7 & 64.5 & 65.4 \\
LLaMA2-13B & Multi & 46.3 & 63.7 & 41.4 & 55.3 \\
Yi-34B & Multi  & 67.1 & 72.2 & 58.7 & \textbf{68.8} \\
\hline
\textbf{Orion-14B} & Multi & \textbf{69.1}  & \textbf{79.5} & \textbf{67.9} & 67.3\\
\hline
\end{tabular}
\end{table}

As shown in Tables \ref{jaeval} and \ref{koeval}, Orion-14 consistently achieves the highest scores across the majority of the test sets. On average, it outperforms all other LLMs in Japanese and Korean datasets, surpassing even those models with a greater number of parameters.

To gain a clearer insight into the multilingual capabilities, we compute the average scores for the evaluation sets in Japanese, Korean, Chinese, and English for comparison. The scores for Japanese and Korean are derived directly from Tables \ref{jaeval} and \ref{koeval}. For the Chinese and English datasets, we calculate the average scores using the OpenCompass dataset, excluding the mathematics and programming testsets.

Table \ref{tab:multilingual} demonstrates Orion-14B's impressive performance in multilingual evaluations. It leads with top scores in Japanese, Korean, and Chinese, surpassing other multilingual models. In English, Orion-14B is marginally outperformed by Yi-34B, which is an LLM with a significantly higher number of parameters. This data highlights Orion-14B's robust proficiency in multiple languages.

\subsection{Analysis of Data Contamination}
\label{subsec:data_contamination}
The rise of the LLM has led to a surge in the performance of evaluation tasks. Their superior performance is primarily attributed to the massive data consumed by these billion/trillion-parameter LLMs during training. However, recent work \citep{yang2023rethinking,golchin2023time,skywork} has shown that the performance of LLM on many downstream tasks may be inflated due to data contamination, i.e., the presence of test data from these downstream tasks in the pretraining data of LLMs. 

As mentioned above, to prevent the pretraining dataset from containing answers to the evaluation datasets, we apply our deduplication approach using all evaluation datasets. This process removes text similar to the evaluation data from the training corpus. On the other hand, to understand the influence of such data, we also experiment with training a model using previously deduplicated data. Specifically, we select those data that had been removed due to deduplication with the evaluation set but we do \emph{not} contain data with the exact same texts as in the evaluation texts. In other words, this approach allows us to use data that may be \emph{semantically} or \emph{partially} related to the evaluation set while excluding the exact text from it. We compile a smaller dataset of 200B tokens, which includes these selected data alongside the regular training data. We then continue the pretraining process with this 200B token dataset, resulting in a new pretrained model named Orion-14B-Exam. As illustrated in the accompanying table, Orion-14B-Exam demonstrates significantly higher scores on the evaluation set compared to the baseline.

\begin{table}[h]
\centering
\caption{Evaluation for data contamination and overfitting.}
\label{tab:data-contam}
\begin{tabular}{l|ccccc}
\hline
Model & C-Eval & CMMLU & MMLU&  Lambada& HellaSwag \\
\hline
GPT-4  & 69.9 & 71 & 83 &  65.5 & \textbf{91.4} \\
Qwen-72B   &83.3  & 61.8 & 77.3 & 76.1 & 85.4 \\
Yi-34B  & 81.8 & 82.6 & 76.3  & 73.1 &  82\\
\hline
Orion-14B  & 72.9 & 70.6 & 69.9 & 78.8 &  78.5\\
Orion-14B-Exam  & \textbf{92.7} & \textbf{82.9} & \textbf{85.4}  & \textbf{78.5} & 85.8 \\
\hline
\end{tabular}
\end{table}

The results in Table \ref{tab:data-contam} reveal that manipulating training data can easily lead to fitting the evaluation dataset and achieve very high scores. We conduct an additional experiment to gauge the extent of overfitting. Specifically, we gather a collection of recent texts from the Internet, ensuring they are unseen in any model's training set. We then calculate the loss on this new dataset $L_{unseen}$ and compare it to the loss on texts drawn from the evaluation sets $L_{eval}$ mentioned in Tables \ref{exp-exam} and \ref{exp-nlu}, including C-Eval, MMLU, HellaSwag, and others. The loss differential between these two sets serves as an indicator of overfitting—the smaller the difference, the lower the likelihood of overfitting to the evaluation set. The results of this analysis are presented in Table \ref{tab:loss}. This table illustrates that with the inclusion of the new training dataset, there is a significant reduction in the loss on the evaluation set, decreasing from 1.87 to 1.44, clearly showing the overfitting on the evaluation set. On the other hand, the original Orion-14B model demonstrates consistent losses on both datasets, with a minimal difference as expected, indicating little overfitting levels.

\begin{table}[h]
\centering
\caption{Overfitting analysis of the loss of each model.}
\label{tab:loss}
\begin{tabular}{l|ccccc}
\hline
Model & $L_{unseen}$ & $L_{eval}$  & $\Delta(L_{unseen}-L_{eval})$\\
\hline
Baichuan2-13B  & 2.23 & 1.93 & 0.30 \\
Qwen-14B & 2.19 & 1.73 & 0.46 \\
Qwen-72B  &2.05 & 1.54 & 0.51 \\
\hline
Orion-14B  & 2.15 & 1.87 & 0.28\\
\hline
Orion-14B-Exam  & 2.18 & 1.44 & 0.74 \\
\hline
\end{tabular}
\end{table}

In light of these performance, it is crucial to examine the evaluation methods used in the community of LLM. Since it is possible to achieve high scores through specific training tactics, such scores may not accurately reflect the true capabilities of an LLM. An overemphasis on top leaderboard positions can be misleading and does not guarantee actual model proficiency. The principal goal should be to develop robust, effective LLMs that prove their utility in a wide range of real-world applications.

\subsection{Fine-tuned Model Evaluations}
The above evaluation utilizes standard evaluation datasets to test the performance of the pretrained foundation model (base-model). On the other hand, evaluating the performance of the fine-tuned model (chat-model) differs from that of the base-model. This is because the chat-model is designed to generate responses to given prompts, and determining the goodness of these responses can be subjective and dependent on the specific task. To comprehensively evaluate the chat-model's performance, we conduct tests using three different approaches: 1) standard evaluation sets, similar to those used in the base-model evaluation; 2) subjective datasets based on GPT-4 scoring; and 3) human evaluation.

\begin{table}[h]
\centering
\caption{Standard evaluation for chat models.}
\label{tab:chat-std}
\begin{tabular}{lccccccl}
\hline
Model &  CMMLU& MMLU&  BBH& HellaSwag&   PIQA&WSC\\
\hline
Baichuan2-13B-Chat  &  58.4 &  57.0 &  49.9&   66.9 &  77.6 & 71.2\\
Qwen-14B-Chat           &  70.0 &  66.4 & 58.0 &   65.2 &  74.0 & 66.3\\
LLaMA2-13B-Chat        &  38.7 &  54.6 & 40.2 &  78.2 &  78.8 & 68.3\\
InternLM-20B-Chat   &   52.2 &  52.5 &  35.3 & 69.2 &  76.7 & 61.5\\
\hline
\textbf{Orion-14B-Chat}         & 63.9 & 61.7 & 49.1\textbf{ }&  76.7 & 78.4 & 71.2\\
\hline
\end{tabular}
\end{table}

For the standard evaluation, we use widely recognized benchmarks, including CMMLU, MMLU, BBH, HellaSwag, PIQA, and WSC. As indicated in \ref{tab:chat-std}, Orion-14B-chat maintains strong performance in HellaSwag, BBH, PIQA, and WSC. However, there is a slight decline in performance on CMMLU and MMLU compared to the base model in Tabels \ref{exp-exam} and \ref{exp-nlu}. This is likely due to the evaluation prompts being more designed for the base model than the chat model. Therefore, incorporating subjective evaluation methods alongside standard metrics could provide a more comprehensive assessment of the model's capabilities. We utilize MT-Bench \citep{zheng2023judging} and AlignBench \citep{liu2023alignbench} for English and Chinese, respectively.

\begin{table}[h]
\centering
\caption{Subjective evaluation of MT-Bench.}
\label{tab:mtbench}
\begin{tabular}{l|cc|c}
\hline
 Model &  First-Turn& Second-Turn &  \textbf{Average}    \\
\hline
Baichuan2-13B-Chat  & 7.05  & 6.47 & 6.76    \\
Qwen-14B-Chat           & 7.30 & 6.62 & 6.96    \\
LLaMA2-13B-Chat        & 7.10 & 6.20 &  6.65    \\
InternLM-20B-Chat    & 7.03 & 5.93 &  6.48   \\
\hline
\textbf{Orion-14B-Chat}         & \textbf{7.68} & \textbf{7.07} & \textbf{ 7.37}    \\
\hline
\end{tabular}
\end{table}

\begin{table}[h]
\centering
\caption{Subjective evaluation of AlignBench. The header of each column stands for Mathematics, Logic, Basic tasks, Chinese understanding, Comprehensive Q\&A, Writing, Role-playing, and Professional tasks, and Average scores.}
\label{tab:alignbench}
\begin{tabular}{l|cccccccc|c}
\hline
Model &Math. & Logi. &  Basic.&  Chi.& Comp.&Writ.& Role.& Prof.& \textbf{Avg.}   \\
\hline
Baichuan2-13B-Chat & 3.76 & 4.07 & 6.22 & 6.05 & 7.11 & 6.97 & 6.75 & 6.43 & 5.25   \\
Qwen-14B-Chat         & \textbf{4.91} & \textbf{4.71}  & \textbf{6.90} & 6.36 & 6.74 & 6.64 & 6.59 & 6.56 & \textbf{5.72}   \\
LLaMA2-13B-Chat      &3.05 & 3.79  & 5.43 & 4.40 & {6.76} & 6.63 & 6.99 & 5.65  & 4.70  \\
InternLM-20B-Chat  & 3.39 & 3.92  & 5.96 & 5.50 & \textbf{7.18} & 6.19 & 6.49 & 6.22 & 4.96  \\
\hline
Orion-14B-Chat    & 4.00 & 4.24  & 6.18 & \textbf{6.57} & 7.16 & \textbf{7.36} & \textbf{7.16} & \textbf{6.99}&  5.51  \\
\hline
\end{tabular}
\end{table}

The results presented in Tables \ref{tab:mtbench} and \ref{tab:alignbench} highlight Orion-14B-Chat's performance in subjective evaluations. In MT-Bench evaluation, Orion-14B-Chat significantly outperforms other models, achieving the highest scores in both First-Turn and Second-Turn evaluations, with an average score of 7.37.
In the AlignBench evaluation, Orion-14B-Chat excels notably in Chinese understanding, Writing, Role-Playing, and Professional tasks. The results demonstrate competitive performance across diverse conversational contexts.

As the chat model is designed to generate responses to prompts, human evaluation is a critical measure of its effectiveness. Adopting an approach similar to the arena method used in Chatbot Arena \citep{lmsys-arena}, we engage human annotators to assess responses from two models in a randomized head-to-head format. Specifically, for a given prompt, responses generated by two anonymized models are presented to the annotators, who then rate them as "Win," "Tie," or "Loss" based on their preference. We have 14 human annotators evaluate a total of 3562 questions. The models compared in this arena battle are Orion-14B-Chat, Qwen-14B-Chat, and Baichuan2-13B-Chat. As indicated in Table \ref{tab:human}, Orion-14B-Chat received the highest number of "win" votes, highlighting its exceptional performance in human evaluations.

\begin{table}[htbp]
\centering
\caption{Human arena evaluation for chat models.}
\label{tab:human}
\begin{tabular}{lccc}
\hline
Model &  Win&  Tie &  Loss  \\
\hline
Orion-14B-Chat & 1172 & 1491 & 899  \\
Qwen-14B-Chat  &1101 &  1592 & 869 \\ 
Baichuan2-13B-Chat & 728 &1601  & 1233    \\
\hline
\end{tabular}
\end{table}

\section{Extension Works}
\label{sec:ext}
In practical applications, LLMs have a variety of needs, including extended context handling, minimizing inference resource requirement, and adapting to specific applications. To address these challenges, we conduct extension works and develop several specialized models. Below are the extensions we have implemented:

\begin{itemize}
\item \textbf{Orion-14B-Long}: This model is optimized for long context lengths more than 200,000 tokens and demonstrates performance comparable to proprietary models on long context evaluation sets \citep{longbench,longeval}.

\item \textbf{Orion-14B-INT4}: A quantized model utilizing 4-bit integer weights. It significantly reduces the model size by 70\% and increases the inference speed by 30\% while incurring a minimal performance loss of only 1\%.

\item \textbf{Orion-14B-RAG}: A chat-model fine-tuned on a custom retrieval augmented generation dataset, achieving superior performance in retrieval augmented generation tasks.

\item \textbf{Orion-14B-PlugIn}: A chat-model specifically tailored for plugin and function calling tasks, ideal for agent-related scenarios where the LLM acts as a plugin and function call system.
\end{itemize}

Due to time constraints, this technical report does not cover the training details and evaluations of these models. We make all the above models available for public use. For more information, please refer to our open-source library {\url{https://github.com/OrionStarAI/Orion}}.

\section{Conclusion}
\label{sec:conclusion}
In this study, we present Orion-14B, a diverse suite of multilingual large language models with 14 billion (14B) parameters. This family includes a pretrained base model and a fine-tuned chat model, as detailed in this technical report. Additionally, we offer several extensions to Orion-14B, such as a long context model, a quantized model, and several application-oriented models, enhancing its versatility and applicability. These models have demonstrated competitive performance against existing open-source models in the field of LLMs, positioning Orion-14B as a potential strong baseline for future LLM research.

Training a large language model like Orion-14B poses considerable challenges. Throughout this endeavor, we faced numerous obstacles and overcame significant hurdles. We responsibly provide open access to the Orion-14B family and documented our experiences and insights in this technical report, aiming to assist and inspire other researchers in the community. 

The journey of LLMs is more than a technological advancement; it is a continuous dialogue between human intelligence and artificial intelligence, constantly evolving and pushing the boundaries of what's possible. As Ludwig Wittgenstein insightfully remarked, "The limits of my language mean the limits of my world." \citep{Wittgenstein} This interplay of language and machine learning does more than just reflect our existing world; it unlocks pathways to previously uncharted realms of understanding.

\bibliographystyle{plainnat}
\bibliography{ref}

\begin{thebibliography}{75}
\providecommand{\natexlab}[1]{#1}
\providecommand{\url}[1]{\texttt{#1}}
\expandafter\ifx\csname urlstyle\endcsname\relax
  \providecommand{\doi}[1]{doi: #1}\else
  \providecommand{\doi}{doi: \begingroup \urlstyle{rm}\Url}\fi

\bibitem[01-ai(2023)]{Yi}
01-ai.
\newblock \url{https://github.com/01-ai/Yi}, 2023.

\bibitem[Bai et~al.(2023{\natexlab{a}})Bai, Bai, Chu, Cui, Dang, Deng, Fan, Ge, Han, Huang, et~al.]{Qwen}
Jinze Bai, Shuai Bai, Yunfei Chu, Zeyu Cui, Kai Dang, Xiaodong Deng, Yang Fan, Wenbin Ge, Yu~Han, Fei Huang, et~al.
\newblock Qwen technical report.
\newblock \emph{arXiv preprint arXiv:2309.16609}, 2023{\natexlab{a}}.

\bibitem[Bai et~al.(2023{\natexlab{b}})Bai, Lv, Zhang, Lyu, Tang, Huang, Du, Liu, Zeng, Hou, Dong, Tang, and Li]{longbench}
Yushi Bai, Xin Lv, Jiajie Zhang, Hongchang Lyu, Jiankai Tang, Zhidian Huang, Zhengxiao Du, Xiao Liu, Aohan Zeng, Lei Hou, Yuxiao Dong, Jie Tang, and Juanzi Li.
\newblock Longbench: A bilingual, multitask benchmark for long context understanding.
\newblock \emph{arXiv preprint arXiv:2308.14508}, 2023{\natexlab{b}}.

\bibitem[Baichuan(2023{\natexlab{a}})]{baichuan1}
Baichuan.
\newblock \url{https://github.com/baichuan-inc/Baichuan-13B}, 2023{\natexlab{a}}.

\bibitem[Baichuan(2023{\natexlab{b}})]{baichuan2}
Baichuan.
\newblock Baichuan 2: Open large-scale language models.
\newblock \emph{arXiv preprint arXiv:2309.10305}, 2023{\natexlab{b}}.
\newblock URL \url{https://arxiv.org/abs/2309.10305}.

\bibitem[Bengio et~al.(2000)Bengio, Ducharme, and Vincent]{bengio2000neural}
Yoshua Bengio, R{\'e}jean Ducharme, and Pascal Vincent.
\newblock A neural probabilistic language model.
\newblock \emph{Advances in neural information processing systems}, 13, 2000.

\bibitem[Bengio et~al.(2009)Bengio, Louradour, Collobert, and Weston]{bengio2009curriculum}
Yoshua Bengio, J{\'e}r{\^o}me Louradour, Ronan Collobert, and Jason Weston.
\newblock Curriculum learning.
\newblock In \emph{Proceedings of the 26th annual international conference on machine learning}, pages 41--48, 2009.

\bibitem[Bisk et~al.(2020)Bisk, Zellers, Gao, Choi, et~al.]{piqa}
Yonatan Bisk, Rowan Zellers, Jianfeng Gao, Yejin Choi, et~al.
\newblock Piqa: Reasoning about physical commonsense in natural language.
\newblock In \emph{Proceedings of the AAAI conference on artificial intelligence}, volume~34, pages 7432--7439, 2020.

\bibitem[Charikar(2002)]{charikar2002similarity}
Moses~S Charikar.
\newblock Similarity estimation techniques from rounding algorithms.
\newblock In \emph{Proceedings of the thiry-fourth annual ACM symposium on Theory of computing}, pages 380--388, 2002.

\bibitem[Chen et~al.(2023)Chen, Roberts, Bhatia, Wang, Zhang, Sala, and R{\'e}]{chen2023skill}
Mayee~F Chen, Nicholas Roberts, Kush Bhatia, Jue Wang, Ce~Zhang, Frederic Sala, and Christopher R{\'e}.
\newblock Skill-it! a data-driven skills framework for understanding and training language models.
\newblock \emph{arXiv preprint arXiv:2307.14430}, 2023.

\bibitem[Christiano et~al.(2017)Christiano, Leike, Brown, Martic, Legg, and Amodei]{christiano2017deep}
Paul~F Christiano, Jan Leike, Tom Brown, Miljan Martic, Shane Legg, and Dario Amodei.
\newblock Deep reinforcement learning from human preferences.
\newblock \emph{Advances in neural information processing systems}, 30, 2017.

\bibitem[Contributors(2023)]{opencompass}
OpenCompass Contributors.
\newblock Opencompass: A universal evaluation platform for foundation models.
\newblock \url{https://github.com/open-compass/opencompass}, 2023.

\bibitem[Dacheng~Li and Zhang(2023)]{longeval}
Anze Xie Ying Sheng Lianmin Zheng Joseph E. Gonzalez Ion Stoica Xuezhe~Ma Dacheng~Li, Rulin~Shao and Hao Zhang.
\newblock How long can open-source llms truly promise on context length?, June 2023.
\newblock URL \url{https://lmsys.org/blog/2023-06-29-longchat}.

\bibitem[Dao(2023)]{dao2023flashattention2}
Tri Dao.
\newblock {FlashAttention-2: Faster Attention with Better Parallelism and Work Partitioning}, 2023.

\bibitem[Devlin et~al.(2019)Devlin, Chang, Lee, and Toutanova]{devlin2019bert}
Jacob Devlin, Ming-Wei Chang, Kenton Lee, and Kristina Toutanova.
\newblock Bert: Pre-training of deep bidirectional transformers for language understanding.
\newblock In \emph{Proceedings of the 2019 Conference of the North American Chapter of the Association for Computational Linguistics: Human Language Technologies}, pages 4171--4186, 2019.

\bibitem[Ding et~al.(2023)Ding, Chen, Xu, Qin, Zheng, Hu, Liu, Sun, and Zhou]{ding2023enhancing}
Ning Ding, Yulin Chen, Bokai Xu, Yujia Qin, Zhi Zheng, Shengding Hu, Zhiyuan Liu, Maosong Sun, and Bowen Zhou.
\newblock Enhancing chat language models by scaling high-quality instructional conversations.
\newblock \emph{arXiv preprint arXiv:2305.14233}, 2023.

\bibitem[Du et~al.(2023)Du, Zong, and Zhang]{du2023mods}
Qianlong Du, Chengqing Zong, and Jiajun Zhang.
\newblock Mods: Model-oriented data selection for instruction tuning, 2023.

\bibitem[Evanson et~al.(2023)Evanson, Lakretz, and King]{evanson2023language}
Linnea Evanson, Yair Lakretz, and Jean-R{\'e}mi King.
\newblock Language acquisition: do children and language models follow similar learning stages?
\newblock \emph{arXiv preprint arXiv:2306.03586}, 2023.

\bibitem[Gao et~al.(2021)Gao, Tow, Biderman, Black, DiPofi, Foster, Golding, Hsu, McDonell, Muennighoff, Phang, Reynolds, Tang, Thite, Wang, Wang, and Zou]{eval-harness}
Leo Gao, Jonathan Tow, Stella Biderman, Sid Black, Anthony DiPofi, Charles Foster, Laurence Golding, Jeffrey Hsu, Kyle McDonell, Niklas Muennighoff, Jason Phang, Laria Reynolds, Eric Tang, Anish Thite, Ben Wang, Kevin Wang, and Andy Zou.
\newblock A framework for few-shot language model evaluation, September 2021.
\newblock URL \url{https://doi.org/10.5281/zenodo.5371628}.

\bibitem[Golchin and Surdeanu(2023)]{golchin2023time}
Shahriar Golchin and Mihai Surdeanu.
\newblock {Time travel in LLMs: Tracing data contamination in large language models}.
\newblock \emph{arXiv preprint arXiv:2308.08493}, 2023.

\bibitem[Hendrycks et~al.(2020)Hendrycks, Burns, Basart, Zou, Mazeika, Song, and Steinhardt]{mmlu}
Dan Hendrycks, Collin Burns, Steven Basart, Andy Zou, Mantas Mazeika, Dawn Song, and Jacob Steinhardt.
\newblock Measuring massive multitask language understanding.
\newblock \emph{arXiv preprint arXiv:2009.03300}, 2020.

\bibitem[Hoffmann et~al.(2022)Hoffmann, Borgeaud, Mensch, Buchatskaya, Cai, Rutherford, Casas, Hendricks, Welbl, Clark, et~al.]{hoffmann2022training}
Jordan Hoffmann, Sebastian Borgeaud, Arthur Mensch, Elena Buchatskaya, Trevor Cai, Eliza Rutherford, Diego de~Las Casas, Lisa~Anne Hendricks, Johannes Welbl, Aidan Clark, et~al.
\newblock Training compute-optimal large language models.
\newblock \emph{arXiv preprint arXiv:2203.15556}, 2022.

\bibitem[Huang et~al.(2023)Huang, Bai, Zhu, Zhang, Zhang, Su, Liu, Lv, Zhang, Lei, et~al.]{Ceval}
Yuzhen Huang, Yuzhuo Bai, Zhihao Zhu, Junlei Zhang, Jinghan Zhang, Tangjun Su, Junteng Liu, Chuancheng Lv, Yikai Zhang, Jiayi Lei, et~al.
\newblock C-eval: A multi-level multi-discipline chinese evaluation suite for foundation models.
\newblock \emph{arXiv preprint arXiv:2305.08322}, 2023.

\bibitem[Indyk and Motwani(1998)]{indyk1998approximate}
Piotr Indyk and Rajeev Motwani.
\newblock Approximate nearest neighbors: towards removing the curse of dimensionality.
\newblock In \emph{Proceedings of the thirtieth annual ACM symposium on Theory of computing}, pages 604--613, 1998.

\bibitem[InternLM(2023)]{2023internlm}
InternLM.
\newblock Internlm: A multilingual language model with progressively enhanced capabilities.
\newblock \url{https://github.com/InternLM/InternLM-techreport}, 2023.

\bibitem[Kaplan et~al.(2020)Kaplan, McCandlish, Henighan, Brown, Chess, Child, Gray, Radford, Wu, and Amodei]{kaplan2020scaling}
Jared Kaplan, Sam McCandlish, Tom Henighan, Tom~B Brown, Benjamin Chess, Rewon Child, Scott Gray, Alec Radford, Jeffrey Wu, and Dario Amodei.
\newblock Scaling laws for neural language models.
\newblock \emph{arXiv preprint arXiv:2001.08361}, 2020.

\bibitem[Kim et~al.(2022)Kim, Jang, Kwon, and Davis]{kobest}
Dohyeong Kim, Myeongjun Jang, Deuk~Sin Kwon, and Eric Davis.
\newblock Kobest: Korean balanced evaluation of significant tasks, 2022.
\newblock URL \url{https://arxiv.org/abs/2204.04541}.

\bibitem[Kim et~al.(2021)Kim, Han, Ham, and Baek]{KoGPT}
Ildoo Kim, Gunsoo Han, Jiyeon Ham, and Woonhyuk Baek.
\newblock Kogpt: Kakaobrain korean(hangul) generative pre-trained transformer.
\newblock \url{https://github.com/kakaobrain/kogpt}, 2021.

\bibitem[Ko et~al.(2023{\natexlab{a}})Ko, Yang, Ryu, Choi, Yang, jiwung Hyun, and Park]{Polyglot-Ko}
Hyunwoong Ko, Kichang Yang, Minho Ryu, Taekyoon Choi, Seungmu Yang, jiwung Hyun, and Sungho Park.
\newblock A technical report for polyglot-ko: Open-source large-scale korean language models, 2023{\natexlab{a}}.

\bibitem[Ko et~al.(2023{\natexlab{b}})Ko, Yang, Ryu, Choi, Yang, Park, et~al.]{ko2023technical}
Hyunwoong Ko, Kichang Yang, Minho Ryu, Taekyoon Choi, Seungmu Yang, Sungho Park, et~al.
\newblock A technical report for polyglot-ko: Open-source large-scale korean language models.
\newblock \emph{arXiv preprint arXiv:2306.02254}, 2023{\natexlab{b}}.

\bibitem[Kojima(2023)]{weblab-10b}
Takeshi Kojima.
\newblock \url{https://huggingface.co/matsuo-lab/weblab-10b}, 2023.

\bibitem[Kudo and Richardson(2018)]{kudo2018sentencepiece}
Taku Kudo and John Richardson.
\newblock Sentencepiece: {A} simple and language independent subword tokenizer and detokenizer for neural text processing.
\newblock \emph{CoRR}, abs/1808.06226, 2018.
\newblock URL \url{http://arxiv.org/abs/1808.06226}.

\bibitem[Lai et~al.(2017)Lai, Xie, Liu, Yang, and Hovy]{lai2017large}
Guokun Lai, Qizhe Xie, Hanxiao Liu, Yiming Yang, and Eduard Hovy.
\newblock Race: Large-scale reading comprehension dataset from examinations.
\newblock \emph{arXiv preprint arXiv:1704.04683}, 2017.

\bibitem[Lee et~al.(2023{\natexlab{a}})Lee, Hunter, and Ruiz]{lee2023platypus}
Ariel~N. Lee, Cole~J. Hunter, and Nataniel Ruiz.
\newblock Platypus: Quick, cheap, and powerful refinement of llms, 2023{\natexlab{a}}.

\bibitem[Lee et~al.(2021)Lee, Ippolito, Nystrom, Zhang, Eck, Callison-Burch, and Carlini]{lee2021deduplicating}
Katherine Lee, Daphne Ippolito, Andrew Nystrom, Chiyuan Zhang, Douglas Eck, Chris Callison-Burch, and Nicholas Carlini.
\newblock Deduplicating training data makes language models better.
\newblock \emph{arXiv preprint arXiv:2107.06499}, 2021.

\bibitem[Lee et~al.(2023{\natexlab{b}})Lee, Nakamura, Shing, McCann, Akiba, and Orii]{JapaneseStableLMBaseAlpha7B}
Meng Lee, Fujiki Nakamura, Makoto Shing, Paul McCann, Takuya Akiba, and Naoki Orii.
\newblock Japanese stablelm base alpha 7b, 2023{\natexlab{b}}.
\newblock URL \url{[https://huggingface.co/stabilityai/japanese-stablelm-base-alpha-7b](https://huggingface.co/stabilityai/japanese-stablelm-base-alpha-7b)}.

\bibitem[Levesque et~al.(2012)Levesque, Davis, and Morgenstern]{WSC}
Hector Levesque, Ernest Davis, and Leora Morgenstern.
\newblock The winograd schema challenge.
\newblock In \emph{Thirteenth International Conference on the Principles of Knowledge Representation and Reasoning}. Citeseer, 2012.

\bibitem[Li et~al.(2023)Li, Zhang, Koto, Yang, Zhao, Gong, Duan, and Baldwin]{cmmlu}
Haonan Li, Yixuan Zhang, Fajri Koto, Yifei Yang, Hai Zhao, Yeyun Gong, Nan Duan, and Timothy Baldwin.
\newblock Cmmlu: Measuring massive multitask language understanding in chinese.
\newblock \emph{arXiv preprint arXiv:2306.09212}, 2023.

\bibitem[Lian et~al.(2023)Lian, Goodson, Pentland, Cook, Vong, and "Teknium"]{OpenOrca}
Wing Lian, Bleys Goodson, Eugene Pentland, Austin Cook, Chanvichet Vong, and "Teknium".
\newblock Openorca: An open dataset of gpt augmented flan reasoning traces.
\newblock \url{https://https://huggingface.co/Open-Orca/OpenOrca}, 2023.

\bibitem[Liu et~al.(2023)Liu, Lei, Wang, Huang, Feng, Wen, Cheng, Ke, Xu, Tam, Zhang, Sun, Wang, Zhang, Huang, Dong, and Tang]{liu2023alignbench}
Xiao Liu, Xuanyu Lei, Shengyuan Wang, Yue Huang, Zhuoer Feng, Bosi Wen, Jiale Cheng, Pei Ke, Yifan Xu, Weng~Lam Tam, Xiaohan Zhang, Lichao Sun, Hongning Wang, Jing Zhang, Minlie Huang, Yuxiao Dong, and Jie Tang.
\newblock Alignbench: Benchmarking chinese alignment of large language models, 2023.

\bibitem[LMSYS(2023)]{lmsys-arena}
LMSYS.
\newblock Chatbot arena leaderboard, 2023.
\newblock URL \url{https://lmsys.org/blog/2023-05-25-leaderboard/}.

\bibitem[Loshchilov and Hutter(2018)]{loshchilov2018fixing}
Ilya Loshchilov and Frank Hutter.
\newblock Fixing weight decay regularization in adam.
\newblock 2018.

\bibitem[Mikolov et~al.(2010)Mikolov, Karafi{\'a}t, Burget, Cernock{\`y}, and Khudanpur]{mikolov2010recurrent}
Tomas Mikolov, Martin Karafi{\'a}t, Lukas Burget, Jan Cernock{\`y}, and Sanjeev Khudanpur.
\newblock Recurrent neural network based language model.
\newblock In \emph{Interspeech}, volume~2, pages 1045--1048. Makuhari, 2010.

\bibitem[Nunes et~al.(2023)Nunes, Heddes, Verg{\'e}s, Abraham, Veidenbaum, Nicolau, and Givargis]{nunes2023dothash}
Igor Nunes, Mike Heddes, Pere Verg{\'e}s, Danny Abraham, Alex Veidenbaum, Alex Nicolau, and Tony Givargis.
\newblock Dothash: Estimating set similarity metrics for link prediction and document deduplication.
\newblock In \emph{Proceedings of the 29th ACM SIGKDD Conference on Knowledge Discovery and Data Mining}, pages 1758--1769, 2023.

\bibitem[NVIDIA(2023)]{apex}
NVIDIA.
\newblock \url{https://github.com/NVIDIA/apex}, 2023.

\bibitem[OpenAI(2022{\natexlab{a}})]{chatgpt}
OpenAI.
\newblock Introducing {ChatGPT}.
\newblock 2022{\natexlab{a}}.

\bibitem[OpenAI(2022{\natexlab{b}})]{gpt4}
OpenAI.
\newblock {GPT-4} technical report.
\newblock \emph{arXiv preprint arXiv:2303.08774}, 2022{\natexlab{b}}.

\bibitem[Ouyang et~al.(2022)Ouyang, Wu, Jiang, Almeida, Wainwright, Mishkin, Zhang, Agarwal, Slama, Ray, et~al.]{ouyang2022training}
Long Ouyang, Jeffrey Wu, Xu~Jiang, Diogo Almeida, Carroll Wainwright, Pamela Mishkin, Chong Zhang, Sandhini Agarwal, Katarina Slama, Alex Ray, et~al.
\newblock Training language models to follow instructions with human feedback.
\newblock \emph{Advances in Neural Information Processing Systems}, 35:\penalty0 27730--27744, 2022.

\bibitem[Paperno et~al.(2016)Paperno, Kruszewski, Lazaridou, Pham, Bernardi, Pezzelle, Baroni, Boleda, and Fernandez]{lambada}
Denis Paperno, Germ\'{a}n Kruszewski, Angeliki Lazaridou, Ngoc~Quan Pham, Raffaella Bernardi, Sandro Pezzelle, Marco Baroni, Gemma Boleda, and Raquel Fernandez.
\newblock The {LAMBADA} dataset: Word prediction requiring a broad discourse context.
\newblock In \emph{Proceedings of the 54th Annual Meeting of the Association for Computational Linguistics (Volume 1: Long Papers)}, pages 1525--1534, Berlin, Germany, August 2016. Association for Computational Linguistics.
\newblock URL \url{http://www.aclweb.org/anthology/P16-1144}.

\bibitem[Penedo et~al.(2023)Penedo, Malartic, Hesslow, Cojocaru, Cappelli, Alobeidli, Pannier, Almazrouei, and Launay]{penedo2023refinedweb}
Guilherme Penedo, Quentin Malartic, Daniel Hesslow, Ruxandra Cojocaru, Alessandro Cappelli, Hamza Alobeidli, Baptiste Pannier, Ebtesam Almazrouei, and Julien Launay.
\newblock The refinedweb dataset for falcon llm: Outperforming curated corpora with web data, and web data only, 2023.

\bibitem[Peters et~al.(2018)Peters, Neumann, Iyyer, Gardner, Clark, Lee, and Zettlemoyer]{Peters:2018}
Matthew~E. Peters, Mark Neumann, Mohit Iyyer, Matt Gardner, Christopher Clark, Kenton Lee, and Luke Zettlemoyer.
\newblock Deep contextualized word representations.
\newblock In \emph{Proceedings of the North American Chapter of the Association for Computational Linguistics (NAACL)}, 2018.

\bibitem[Preferred~Networks(2023)]{PLaMo2023Introducing}
Inc Preferred~Networks.
\newblock Plamo-13b, 2023.
\newblock URL \url{https://huggingface.co/pfnet/plamo-13b}.

\bibitem[Radford et~al.(2018)Radford, Narasimhan, Salimans, Sutskever, et~al.]{radford2018improving}
Alec Radford, Karthik Narasimhan, Tim Salimans, Ilya Sutskever, et~al.
\newblock Improving language understanding by generative pre-training.
\newblock 2018.

\bibitem[Rafailov et~al.(2023)Rafailov, Sharma, Mitchell, Ermon, Manning, and Finn]{rafailov2023direct}
Rafael Rafailov, Archit Sharma, Eric Mitchell, Stefano Ermon, Christopher~D Manning, and Chelsea Finn.
\newblock Direct preference optimization: Your language model is secretly a reward model.
\newblock \emph{arXiv preprint arXiv:2305.18290}, 2023.

\bibitem[Sasaki et~al.(2023)Sasaki, Hirakawa, Horie, and Nakamura]{elyzaLLaMA2023}
Akira Sasaki, Masato Hirakawa, Shintaro Horie, and Tomoaki Nakamura.
\newblock Elyza-japanese-llama-2-7b, 2023.
\newblock URL \url{https://huggingface.co/elyza/ELYZA-japanese-LLaMA-2-7b}.

\bibitem[Shibata et~al.(1999)Shibata, Kida, Fukamachi, Takeda, Shinohara, Shinohara, and Arikawa]{shibata1999byte}
Yusuxke Shibata, Takuya Kida, Shuichi Fukamachi, Masayuki Takeda, Ayumi Shinohara, Takeshi Shinohara, and Setsuo Arikawa.
\newblock Byte pair encoding: A text compression scheme that accelerates pattern matching.
\newblock 1999.

\bibitem[Shoeybi et~al.(2020)Shoeybi, Patwary, Puri, LeGresley, Casper, and Catanzaro]{shoeybi2020megatronlm}
Mohammad Shoeybi, Mostofa Patwary, Raul Puri, Patrick LeGresley, Jared Casper, and Bryan Catanzaro.
\newblock Megatron-lm: Training multi-billion parameter language models using model parallelism, 2020.

\bibitem[Su et~al.(2021)Su, Lu, Pan, Murtadha, Wen, and Liu]{su2021roformer}
Jianlin Su, Yu~Lu, Shengfeng Pan, Ahmed Murtadha, Bo~Wen, and Yunfeng Liu.
\newblock Roformer: Enhanced transformer with rotary position embedding.
\newblock \emph{arXiv preprint arXiv:2104.09864}, 2021.

\bibitem[Suzgun et~al.(2022)Suzgun, Scales, Sch{\"a}rli, Gehrmann, Tay, Chung, Chowdhery, Le, Chi, Zhou, , and Wei]{suzgun2022challenging}
Mirac Suzgun, Nathan Scales, Nathanael Sch{\"a}rli, Sebastian Gehrmann, Yi~Tay, Hyung~Won Chung, Aakanksha Chowdhery, Quoc~V Le, Ed~H Chi, Denny Zhou, , and Jason Wei.
\newblock Challenging big-bench tasks and whether chain-of-thought can solve them.
\newblock \emph{arXiv preprint arXiv:2210.09261}, 2022.

\bibitem[THUDM(2023)]{Chatglm3}
THUDM.
\newblock \url{https://github.com/THUDM/ChatGLM3}, 2023.

\bibitem[Touvron et~al.(2023{\natexlab{a}})Touvron, Lavril, Izacard, Martinet, Lachaux, Lacroix, Rozi{\`e}re, Goyal, Hambro, Azhar, et~al.]{LLaMA}
Hugo Touvron, Thibaut Lavril, Gautier Izacard, Xavier Martinet, Marie-Anne Lachaux, Timoth{\'e}e Lacroix, Baptiste Rozi{\`e}re, Naman Goyal, Eric Hambro, Faisal Azhar, et~al.
\newblock Llama: Open and efficient foundation language models.
\newblock \emph{arXiv preprint arXiv:2302.13971}, 2023{\natexlab{a}}.

\bibitem[Touvron et~al.(2023{\natexlab{b}})Touvron, Martin, Stone, Albert, Almahairi, Babaei, Bashlykov, Batra, Bhargava, Bhosale, et~al.]{LLaMA2}
Hugo Touvron, Louis Martin, Kevin Stone, Peter Albert, Amjad Almahairi, Yasmine Babaei, Nikolay Bashlykov, Soumya Batra, Prajjwal Bhargava, Shruti Bhosale, et~al.
\newblock Llama 2: Open foundation and fine-tuned chat models.
\newblock \emph{arXiv preprint arXiv:2307.09288}, 2023{\natexlab{b}}.

\bibitem[Vaswani et~al.(2017)Vaswani, Shazeer, Parmar, Uszkoreit, Jones, Gomez, Kaiser, and Polosukhin]{vaswani2017attention}
Ashish Vaswani, Noam Shazeer, Niki Parmar, Jakob Uszkoreit, Llion Jones, Aidan~N Gomez, {\L}ukasz Kaiser, and Illia Polosukhin.
\newblock Attention is all you need.
\newblock In \emph{Proceedings of the Conference on Neural Information Processing Systems (NIPS 2017)}, pages 5998--6008, 2017.

\bibitem[Wei et~al.(2023)Wei, Zhao, Zhang, Zhu, Wang, Yang, Li, Cheng, L{\"u}, Hu, et~al.]{skywork}
Tianwen Wei, Liang Zhao, Lichang Zhang, Bo~Zhu, Lijie Wang, Haihua Yang, Biye Li, Cheng Cheng, Weiwei L{\"u}, Rui Hu, et~al.
\newblock Skywork: A more open bilingual foundation model.
\newblock \emph{arXiv preprint arXiv:2310.19341}, 2023.

\bibitem[Wenting~Zhao(2023)]{zhao2023wildchat}
Jack Hessel Claire Cardie Yejin Choi Yuntian~Deng. Wenting~Zhao, Xiang~Ren.
\newblock (inthe)wildchat: 650k chatgpt interaction logs in the wild, 2023.

\bibitem[Wittgenstein(1922)]{Wittgenstein}
Ludwig Wittgenstein.
\newblock \emph{Tractatus logigo-philosphicus}.
\newblock 1922.

\bibitem[Yang et~al.(2023)Yang, Chiang, Zheng, Gonzalez, and Stoica]{yang2023rethinking}
Shuo Yang, Wei-Lin Chiang, Lianmin Zheng, Joseph~E Gonzalez, and Ion Stoica.
\newblock Rethinking benchmark and contamination for language models with rephrased samples.
\newblock \emph{arXiv preprint arXiv:2311.04850}, 2023.

\bibitem[Yao et~al.(2013)Yao, Zweig, Hwang, Shi, and Yu]{yao2013recurrent}
Kaisheng Yao, Geoffrey Zweig, Mei-Yuh Hwang, Yangyang Shi, and Dong Yu.
\newblock Recurrent neural networks for language understanding.
\newblock In \emph{Interspeech}, pages 2524--2528, 2013.

\bibitem[Yuanxiang(2023)]{XVERSE-13B}
Yuanxiang.
\newblock \url{https://github.com/xverse-ai/XVERSE-13B}, 2023.

\bibitem[Zellers et~al.(2019)Zellers, Holtzman, Bisk, Farhadi, and Choi]{hellaswag}
Rowan Zellers, Ari Holtzman, Yonatan Bisk, Ali Farhadi, and Yejin Choi.
\newblock Hellaswag: Can a machine really finish your sentence?
\newblock \emph{arXiv preprint arXiv:1905.07830}, 2019.

\bibitem[Zhang et~al.(2023{\natexlab{a}})Zhang, Shi, Liu, Yuan, Li, Dong, Shu, Li, Wang, Lin, Huang, and Fu]{zhang2023chinese}
Ge~Zhang, Yemin Shi, Ruibo Liu, Ruibin Yuan, Yizhi Li, Siwei Dong, Yu~Shu, Zhaoqun Li, Zekun Wang, Chenghua Lin, Wenhao Huang, and Jie Fu.
\newblock Chinese open instruction generalist: A preliminary release, 2023{\natexlab{a}}.

\bibitem[Zhang et~al.(2023{\natexlab{b}})Zhang, Li, Zong, Ying, He, and Qiu]{Zhang2023EvaluatingTP}
Xiaotian Zhang, Chunyang Li, Yi~Zong, Zhengyu Ying, Liang He, and Xipeng Qiu.
\newblock Evaluating the performance of large language models on gaokao benchmark.
\newblock 2023{\natexlab{b}}.

\bibitem[Zheng et~al.(2023)Zheng, Chiang, Sheng, Zhuang, Wu, Zhuang, Lin, Li, Li, Xing, Zhang, Gonzalez, and Stoica]{zheng2023judging}
Lianmin Zheng, Wei-Lin Chiang, Ying Sheng, Siyuan Zhuang, Zhanghao Wu, Yonghao Zhuang, Zi~Lin, Zhuohan Li, Dacheng Li, Eric~P. Xing, Hao Zhang, Joseph~E. Gonzalez, and Ion Stoica.
\newblock {Judging LLM-as-a-Judge with MT-Bench and Chatbot Arena}, 2023.

\bibitem[Zhong et~al.(2023)Zhong, Cui, Guo, Liang, Lu, Wang, Saied, Chen, and Duan]{zhong2023agieval}
Wanjun Zhong, Ruixiang Cui, Yiduo Guo, Yaobo Liang, Shuai Lu, Yanlin Wang, Amin Saied, Weizhu Chen, and Nan Duan.
\newblock Agieval: A human-centric benchmark for evaluating foundation models, 2023.

\bibitem[Zhou et~al.(2023)Zhou, Liu, Xu, Iyer, Sun, Mao, Ma, Efrat, Yu, Yu, Zhang, Ghosh, Lewis, Zettlemoyer, and Levy]{zhou2023lima}
Chunting Zhou, Pengfei Liu, Puxin Xu, Srini Iyer, Jiao Sun, Yuning Mao, Xuezhe Ma, Avia Efrat, Ping Yu, Lili Yu, Susan Zhang, Gargi Ghosh, Mike Lewis, Luke Zettlemoyer, and Omer Levy.
\newblock Lima: Less is more for alignment, 2023.

\end{thebibliography}

\newpage

\appendix
\addcontentsline{toc}{section}{Appendix}
\section*{Appendix}

\section{Contributions}
\label{sec:authors}
All contributors sorted alphabetically by last name.

Core Contributors: Du Chen, Yi Huang, Xiaopu Li, Yongqiang Li, Yongqiang Liu, Haihui Pan, Leichao Xu, Dacheng Zhang, Zhipeng Zhang.

Contributors: Yang Fan, Xuefeng Li, Yuxiang Liu, Haonan Tan, Bingcheng Zhang, Enmao Zhang, Yinglou Zhao.

Human Annotators: Lixiu Chen, Zhenwei Hu, Ningting Luo, Zikang Ma, Jiali Pan, Yuping Qin, Qin Shu, Qin Tu, Haiyan Wu, Jiamin Wu, Jingping Wu, Jing Xia, Simiao Xu, Zhiyong Xue, Chonghuan Yang, Tao Zhu.

Science and Engineering Leadership: Kun Han.

We thank the executive team for their support: Sheng Fu, Mingyan Sun, Ting Li.

\section{Detailed evaluation results of OpenCompass}
\label{sec:opencompass}

\begin{table}[h]
\centering
\caption{Evaluation results of OpenCompass in the examination category}
\label{tab:oc2}
\begin{tabular}{l|c|ccccccc}
\hline
Model & Average & C-Eval & CMMLU & MMLU & AGIEval  &  GaoKao & ARC-c & ARC-e\\
\hline
LLaMA 2-13B & 45.2 &  41.4&  38.4&  55.0&  30.9&  18.2&  60.3&  71.8\\
Skywork-13B & 61.1 &  59.1&  61.4&  62.7&  43.6&  56.1&  65.4&  79.5\\
Baichuan 2-13B & 51.8 &  59.0&  61.3&  59.5&  37.4&  45.6&  38 &  61.9\\
Qwen-14B & 71.3 &  71.7&  70.2&  67.9&  51.9&  \textbf{62.5}& \textbf{84.4} & \textbf{90.1}\\
InternLM-20B & 62.5 &  58.8&  59.0&  62.1&  44.6&  45.5&  81.7 &  86.1\\
\hline
\textbf{Orion-14B} & \textbf{71.4} &  \textbf{72.9}&  \textbf{70.6}&  \textbf{69.9}& \textbf{54.7}&  62.1&  80.7  &  88.9\\
\hline
\end{tabular}
\end{table}

\begin{table}[h]
\centering
\caption{Evaluation results of OpenCompass in the language category}
\label{tab:oc2}
\begin{tabular}{l|c|cccccc}
\hline
Model & Average  & WiC & CHID & AFQMC & WSC  &  TyDiQA & Flores \\
\hline
LLaMA 2-13B & 47.0 &  53.3&  53.0&  69.0&  66.3&  33.2 &  7.20\\
Skywork-13B & 51.3 &  51.1&  88.1&  69.0&  66.3&  27.9 &  5.40\\
Baichuan 2-13B & 47.5 &  60.2&  83.2&  38.0&  66.3&  30.8 &  6.40\\
Qwen-14B & 52.7& 50.9 &  84.7&  69.0&  66.3&  39.8 &  5.30\\
InternLM-20B & 55.0 &  \textbf{61.8} &  81.7&  69.0&  68.3&  \textbf{43.2} &  6.00\\
\hline
\textbf{Orion-14B}  & \textbf{55.0} &  60.0&  \textbf{90.1}&  \textbf{69.0}&  \textbf{70.2}&  32.7  &  \textbf{8.13}\\
\hline
\end{tabular}
\end{table}

\begin{table}[h]
\centering
\caption{Evaluation results of OpenCompass in the knowledge category}
\label{tab:oc2}
\begin{tabular}{l|c|cccc}
\hline
Model & Average & BoolQ & CommonSenseQA & TriviaQA & NaturalQuestions \\
\hline
LLaMA 2-13B & 58.3 &  82.4&  66.7&  59.4&  24.8\\
Skywork-13B & 52.7 &  80.9&  64.6&  48.1&  17.2\\
Baichuan 2-13B & 48.9 &  67&  65.6&  46.6&  16.3\\
Qwen-14B & 56.1 &  86.1&  70.1&  48.4&  19.8\\
InternLM-20B & \textbf{60.1} & \textbf{87.5}& \textbf{70.6}&  57.3& \textbf{25.2}\\
\hline
\textbf{Orion-14B} & 60.0 &  84.9&  65.7& \textbf{77.2}&  12.4\\
\hline
\end{tabular}
\end{table}

\begin{table}[h]
\centering
\caption{Evaluation results of OpenCompass in the understanding category}
\label{tab:oc2}
\begin{tabular}{l|c|cccc}
\hline
Model & Average & C3 &  RACE-middle & RACE-high & OpenbookQA \\
\hline
LLaMA 2-13B  & 50.9 &  46.1&  63.0&  58.9&  65.0\\
Skywork-13B  & 64.5&  64.9&  87.6&  84.1&  83.4\\
Baichuan 2-13B & 58.1&  65.6 &  68.9&  67.2&  65.0\\
Qwen-14B & 68.8& \textbf{90.8} &  93.0&  90.3& \textbf{94.8}\\
InternLM-20B & 67.3&  73.7 &  86.4&  83.3&  87.6\\
\hline
\textbf{Orion-14B} & \textbf{71.9} &  80.2  & \textbf{93.2}& \textbf{91.3}&  89.8\\
\hline
\end{tabular}
\end{table}

\begin{table}[h]
\centering
\caption{Evaluation results of OpenCompass in the understanding category (cont')}
\label{tab:oc2}
\begin{tabular}{l|ccccc}
\hline
Model & CSL & LCSTS & XSum & EPRSTMT  &  Lambada \\
\hline
LLaMA 2-13B  &  58.8&  7.80&  23.4 &  58.8&  76.5\\
Skywork-13B  &  60.0&  17.7&  22.6 &  88.1&  71.8\\
Baichuan 2-13B &  63.1&  6.30&  25.2 &  87.5&  74.1\\
Qwen-14B &  54.4&  12.5&  24.7 &  86.9&  71.4\\
InternLM-20B & \textbf{65.6}&  12.7&  35.5 & \textbf{89.4}&  71.8\\
\hline
\textbf{Orion-14B} &  62.5& \textbf{28.9}& \textbf{38.2} &  83.8& \textbf{78.8}\\
\hline
\end{tabular}
\end{table}

\begin{table}[h]
\centering
\caption{Evaluation results of OpenCompass in the reasoning category}
\label{tab:oc2}
\begin{tabular}{l|c|ccccccccc}
\hline
Model & Average & CMNLI & OCNLI & AXb & AXg & RTE & COPA  &  ReCoRD \\
\hline
LLaMA 2-13B & 43.6 &  41.4&  34.1&  58.3&  50.6 &  47.3&  70.0&  11.6\\
Skywork-13B & 45.2 &  32.8&  30.0&  59.0&  53.4 &  56.3&  72.0&  1.40\\
Baichuan 2-13B & 44.2 &  32.7&  30.0&  59.7&  50.6 &  44.8&  71.0&  20.7\\
Qwen-14B & 60.1 &  62.1&  58.2&  49.5&  80.9 &  71.5&  \textbf{93.0}&  42.3\\
InternLM-20B & 54.9 &  43.0&  42.5&  62.1&  75.0 &  57.8&  83.0&  63.6\\
\hline
\textbf{Orion-14B} & \textbf{61.6} & \textbf{72.6}& \textbf{68.3}& \textbf{71.2}& \textbf{86.5} & \textbf{83.0}&  82.0&  \textbf{87.8}\\
\hline
\end{tabular}
\end{table}

\begin{table}[h]
\centering
\caption{Evaluation results of OpenCompass in the reasoning category (cont')}
\label{tab:oc2}
\begin{tabular}{l|ccccccccc}
\hline
Model & HellaSwag & PIQA & SIQA & MATH & GSM8K & DROP  &  HumanEval & MBPP & BBH \\
\hline
LLaMA 2-13B&  77.5&  79.8&  54.8 &  5.00&  29.6&  46.4&  18.9&  26.8&  45.6\\
Skywork-13B&  73.7&  78.3&  70.4 &  9.80&  54.3&  41.7&  15.9&  25.4&  48.3\\
Baichuan 2-13B&  70.8&  78.1&  44.3 &  10.1&  52.6&  45.0&  17.1&  30.8&  49.0\\
Qwen-14B&  \textbf{80.2}&  79.8&  \textbf{78.1} & \textbf{25.2}& \textbf{61.6}& \textbf{53.6}& \textbf{32.3}&  \textbf{39.8}&  53.7\\
InternLM-20B&  78.1& \textbf{80.3}&  72.8 &  7.90&  52.6&  46.0&  25.6&  35.6&  52.5\\
\hline
\textbf{Orion-14B}&  78.5&  79.5&  69.4  &  7.78&  51.9&  40.8&  20.7&  29.0& \textbf{56.5}\\
\hline
\end{tabular}
\end{table}

\end{document}